\documentclass[10pt,twocolumn,letterpaper]{article}

\usepackage[noblocks]{authblk}
\usepackage[pagenumbers]{cvpr}

\usepackage{multirow}
\usepackage{colortbl}

\definecolor{ceiling}{RGB}{214, 38, 40}
\definecolor{floor}{RGB}{43, 160, 4}
\definecolor{wall}{RGB}{158, 216, 229}
\definecolor{window}{RGB}{114, 158, 206}
\definecolor{chair}{RGB}{204, 204, 91}
\definecolor{bed}{RGB}{255, 186, 119}
\definecolor{sofa}{RGB}{147, 102, 188}
\definecolor{table}{RGB}{30, 119, 181}
\definecolor{tvs}{RGB}{160, 188, 33}
\definecolor{furniture}{RGB}{255, 127, 12}
\definecolor{objects}{RGB}{196, 175, 214}

\definecolor{nbarrier}{RGB}{255, 120, 50}
\definecolor{nbicycle}{RGB}{255, 192, 203}
\definecolor{nbus}{RGB}{255, 255, 0}
\definecolor{ncar}{RGB}{0, 150, 245}
\definecolor{nconstruct}{RGB}{0, 255, 255}
\definecolor{nmotor}{RGB}{200, 180, 0}
\definecolor{npedestrian}{RGB}{255, 0, 0}
\definecolor{ntraffic}{RGB}{255, 240, 150}
\definecolor{ntrailer}{RGB}{135, 60, 0}
\definecolor{ntruck}{RGB}{160, 32, 240}
\definecolor{ndriveable}{RGB}{255, 0, 255}
\definecolor{nother}{RGB}{139, 137, 137}
\definecolor{nsidewalk}{RGB}{75, 0, 75}
\definecolor{nterrain}{RGB}{150, 240, 80}
\definecolor{nmanmade}{RGB}{213, 213, 213}
\definecolor{nvegetation}{RGB}{0, 175, 0}

\definecolor{nvcolor}{RGB}{119, 185, 0}
\definecolor{roadcolor}{RGB}{234, 51, 246}
\definecolor{sidewalkcolor}{RGB}{68, 8, 72}
\definecolor{parkingcolor}{RGB}{241, 156, 249}
\definecolor{othergroundcolor}{RGB}{160, 32, 76}
\definecolor{buildingcolor}{RGB}{246, 202, 69}
\definecolor{carcolor}{RGB}{111, 149, 238}
\definecolor{truckcolor}{RGB}{74, 32, 172}
\definecolor{bicyclecolor}{RGB}{136, 227, 242}
\definecolor{motorcyclecolor}{RGB}{37, 59, 146}
\definecolor{othervehiclecolor}{RGB}{96, 81, 242}
\definecolor{vegetationcolor}{RGB}{79, 173, 50}
\definecolor{trunkcolor}{RGB}{126, 65, 22}
\definecolor{terraincolor}{RGB}{171, 238, 105}
\definecolor{personcolor}{RGB}{234, 60, 49}
\definecolor{bicyclistcolor}{RGB}{234, 66, 195}
\definecolor{motorcyclistcolor}{RGB}{138, 42, 90}
\definecolor{fencecolor}{RGB}{238, 128, 69}
\definecolor{polecolor}{RGB}{252, 241, 161}
\definecolor{trafficsigncolor}{RGB}{233, 51, 35}
\definecolor{other-struct.color}{RGB}{255, 150, 0}
\definecolor{other-objectcolor}{RGB}{50, 255, 255}
\definecolor{lane-markingcolor}{RGB}{150, 255, 170}
\definecolor{color1}{RGB}{176, 36, 24}
\definecolor{color2}{RGB}{0, 176, 80}
\definecolor{color3}{RGB}{0, 0, 200}

\definecolor{cvprblue}{rgb}{0.21,0.49,0.74}
\usepackage[breaklinks,colorlinks,allcolors=cvprblue]{hyperref}

\usepackage[authorsmultiline]{latexml}

\title{OccAnyScene: Towards Unified Indoor-Outdoor 3D Occupancy Prediction}
\iflatexml
  \author[1,4]{Junjie Liu\textsuperscript{*}}
  \author[2]{Wanshui Gan\textsuperscript{*}}
  \author[3]{Zitong Dai}
  \author[1]{Guiping Cao}
  \author[1]{Yan Li}
  \author[1]{Ke Chen}
  \author[1]{Dongmei Jiang}
  \author[4]{Jianguo Zhang}
  \author[1]{Xiangyuan Lan\textsuperscript{\textdagger}}
  \affil[1]{Pengcheng Laboratory}
  \affil[2]{Shanghai Artificial Intelligence Laboratory}
  \affil[3]{HITSZ}
  \affil[4]{SUSTech}
  \date{\small
    \textsuperscript{*} Equal contribution.\quad
    \textsuperscript{\textdagger} Corresponding author.\\
    \href{https://RoboPerception.github.io/OccAnyScene/}{Project Page}
    \quad $\vert$ \quad
    \href{https://github.com/RoboPerception/OccAnyScene}{Code Repository}}
\else
  \author{\begin{tabular}{c}
    Junjie Liu\textsuperscript{*,1,4}\quad
    Wanshui Gan\textsuperscript{*,2}\quad
    Zitong Dai\textsuperscript{3}\quad
    Guiping Cao\textsuperscript{1}\authorcr
    Yan Li\textsuperscript{1}\quad
    Ke Chen\textsuperscript{1}\quad
    Dongmei Jiang\textsuperscript{1}\quad
    Jianguo Zhang\textsuperscript{4}\quad
    Xiangyuan Lan\textsuperscript{\textdagger,1}\\
    {\scriptsize
      \textsuperscript{1}Pengcheng Laboratory\quad
      \textsuperscript{2}Shanghai Artificial Intelligence Laboratory\quad
      \textsuperscript{3}HITSZ\quad
      \textsuperscript{4}SUSTech}\\
    {\scriptsize
      \textsuperscript{*} Equal contribution.\quad
      \textsuperscript{\textdagger} Corresponding author.}\\
    {\scriptsize
      \href{https://RoboPerception.github.io/OccAnyScene/}{Project Page}
      \quad $\vert$ \quad
      \href{https://github.com/RoboPerception/OccAnyScene}{Code Repository}}
  \end{tabular}}
  \date{}
\fi

\begin{document}

\maketitle
\iflatexml\else
  \vspace{-1.3em}
\fi

\begin{abstract}
3D occupancy prediction is fundamental to scene understanding, yet existing 3D semantic occupancy methods are typically specialized to fixed scene types and occupancy protocols. We introduce \textbf{Cross-Scene 3D Semantic Occupancy Prediction}, a new task setting which requires a single model to handle heterogeneous indoor and outdoor scenes with varying cameras, spatial ranges, voxel specifications, and semantic taxonomies. This setting poses a fundamental challenge: achieving metric-consistent yet scene-adaptive image-to-3D lifting across varying camera configurations and scene scales. To address this challenge, we propose \textbf{OccAnyScene}, a pixel-frustum-centered Gaussian framework built upon a pretrained depth foundation model. Specifically, the framework employs \textbf{Pixel-Aligned Frustum Feature Aggregation} to construct a camera-aware frustum query for each feature pixel, and \textbf{Frustum-Parameterized Gaussian Construction} to decode each query into multiple Gaussians whose positions and sizes are constrained by the predicted pixel depth and corresponding frustum geometry. OccAnyScene sets new state-of-the-art results, achieving \(59.92\%\) mIoU on  the indoor Occ-ScanNet and \(23.06\%\) mIoU on the outdoor SurroundOcc-nuScenes.
\end{abstract}

\section{Introduction}

\begin{figure}[!t]
\centering
\includegraphics[width=1\columnwidth]{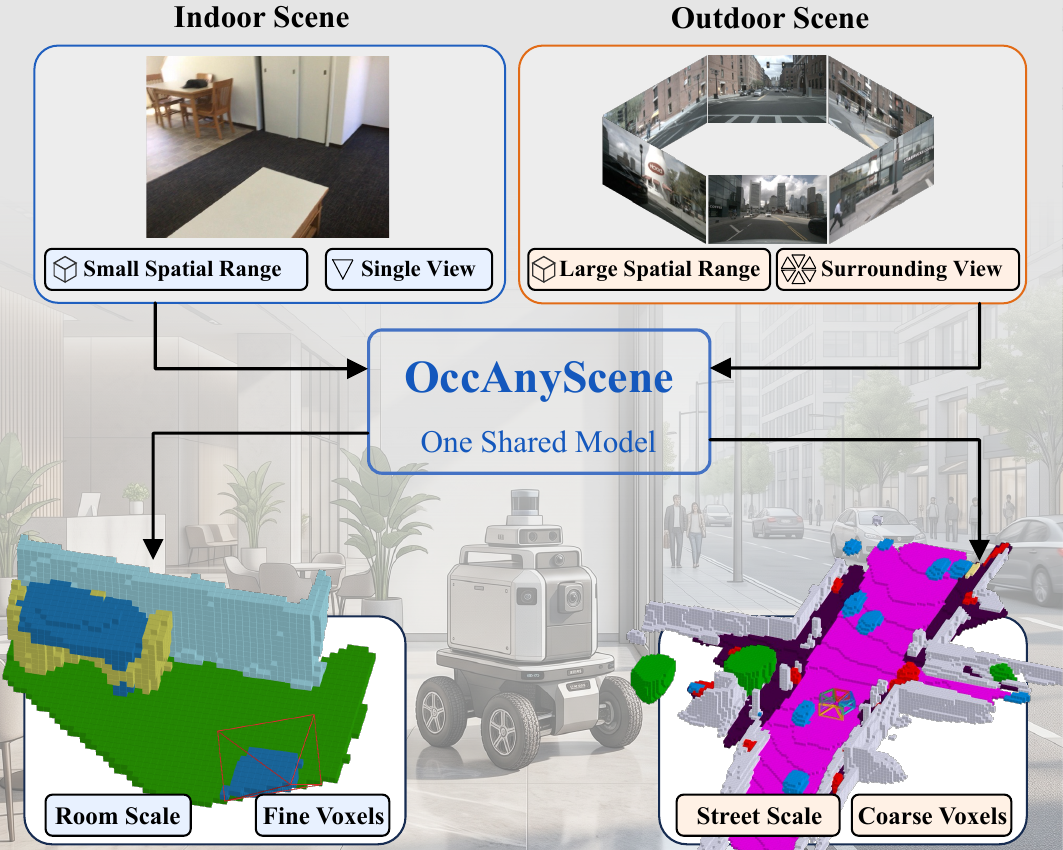}
\caption{
Cross-scene 3D semantic occupancy prediction with OccAnyScene.
OccAnyScene jointly learns a single model for heterogeneous indoor and outdoor scenes.
}
\label{fig:teaser}
\end{figure}

3D semantic occupancy prediction is a fundamental task for holistic 3D scene understanding, with broad applications in autonomous driving~\cite{ma2024vision,hayes20253d} and embodied AI~\cite{qin2025robofactory,wang2024embodiedscan,wu2025embodiedocc}. It reconstructs dense geometric and semantic states over both visible and occluded regions, providing essential scene information for planning, navigation, and interaction. However, existing approaches are predominantly developed as scene-specific occupancy models, each designed for a particular indoor scene type or outdoor driving setting. Their spatial ranges, voxel resolutions, camera configurations, and semantic prediction spaces are therefore tailored to the corresponding dataset protocols~\cite{huang2023tri,zhang2023occformer,yu2024monocular}. This specialization limits the development of occupancy prediction toward general-purpose 3D perception.

Such generality is necessary for autonomous systems operating across environments with different spatial demands. For example, an autonomous vehicle requires long-range but coarse-grained occupancy perception on open roads, yet short-range and fine-grained perception in confined indoor parking spaces. Maintaining and switching between separate models for these settings complicates deployment and scales poorly as operating environments diversify. We therefore introduce \textbf{Cross-Scene 3D Semantic Occupancy Prediction}, where one model is jointly trained across heterogeneous indoor and outdoor scenes under their native sensing and output protocols, as illustrated in Fig.~\ref{fig:teaser}. This setting requires the model to learn a shared scene representation while adapting to different spatial and semantic definitions.

The difficulty of cross-scene occupancy goes beyond the visual domain gap between indoor and outdoor scenes. Existing occupancy models typically encode a fixed spatial support into their scene representations, which becomes restrictive across domains with different scene extents and voxel resolutions. Dense voxel representations~\cite{li2023voxformer,liu2024fully} are inherently tied to predefined spatial ranges and resolutions. Continuous 3D Gaussian primitives ~\cite{kerbl3Dgaussians} offer greater flexibility, while depth-based methods~\cite{huang2025gaussianformer,qian2026splatssc} adapt Gaussian centers to observed geometry by lifting pixels according to predicted depth, but initially cover only visible surfaces. Existing methods extend them into occluded regions either through learned position offsets~\cite{szymanowicz2024splatter,huang2025gaussianformer} or by placing multiple Gaussians along each feature pixel ray~\cite{szymanowicz2025flash3d,huang2025gaussianformer}. Under a scene-specific protocol, Gaussian offsets and scales follow relatively consistent metric distributions. Across indoor and outdoor scenes, however, the offsets required for occlusion completion span substantially different metric ranges. Moreover, the appropriate Gaussian scale depends on depth and camera intrinsics: neighboring feature pixels become farther apart in 3D after back-projection at greater depths, requiring larger Gaussians to maintain spatial coverage. These differences make a shared absolute parameterization difficult to learn. Consequently, stable cross-scene Gaussian construction requires a geometric reference that adapts Gaussian positions and scales to different camera geometries and scene ranges.

To address this challenge, we propose \textbf{OccAnyScene}, a pixel-frustum-centered Gaussian framework for cross-scene occupancy prediction. Our key insight is that each feature pixel corresponds to a finite camera-dependent 3D frustum rather than only its central ray. At a predicted depth, the frustum geometry specifies the metric spatial extent represented by that pixel, which automatically varies with camera intrinsics and scene depth. Based on this insight, OccAnyScene extends a pretrained depth foundation model with a simple yet effective two-module design for constructing the pixel-frustum-centered representation.
\textbf{Pixel-Aligned Frustum Feature Aggregation (PFFA)} conditions each pixel-aligned geometric feature on its camera ray and aggregates surrounding context through cross-attention to reason about occluded regions, producing a frustum query for each feature pixel. \textbf{Frustum-Parameterized Gaussian Construction (FPGC)} converts the frustum queries into Gaussian primitives, using predicted visible surfaces as depth anchors and learning frustum-constrained center offsets to extend the representation into occluded regions, while adapting Gaussian scales to the depth-dependent frustum geometry. Finally, the Gaussian representation is mapped to scene-specific semantic categories and splatted onto the target voxel grid.

\begin{itemize}
    \item We introduce \textbf{Cross-Scene 3D Semantic Occupancy Prediction}, a new task setting that aims to construct a single model for heterogeneous indoor and outdoor scenes with different camera configurations, spatial ranges, voxel specifications, and semantic taxonomies.
    
    \item We propose \textbf{OccAnyScene}, a pixel-frustum-centered Gaussian framework that unifies indoor and outdoor occupancy prediction in one model through \textbf{PFFA} for frustum representation and \textbf{FPGC} for scene-adaptive Gaussian construction across visible and occluded regions.
    
    \item Extensive experiments demonstrate that OccAnyScene achieves state-of-the-art performance on the indoor Occ-ScanNet and outdoor SurroundOcc-nuScenes benchmarks. Moreover, the cross-scene model performs comparably to separately trained scene-specific models, demonstrating its effectiveness across heterogeneous scenes.
\end{itemize}

\section{Related Works}

\begin{figure*}[t]
\centering
\includegraphics[width=\textwidth]{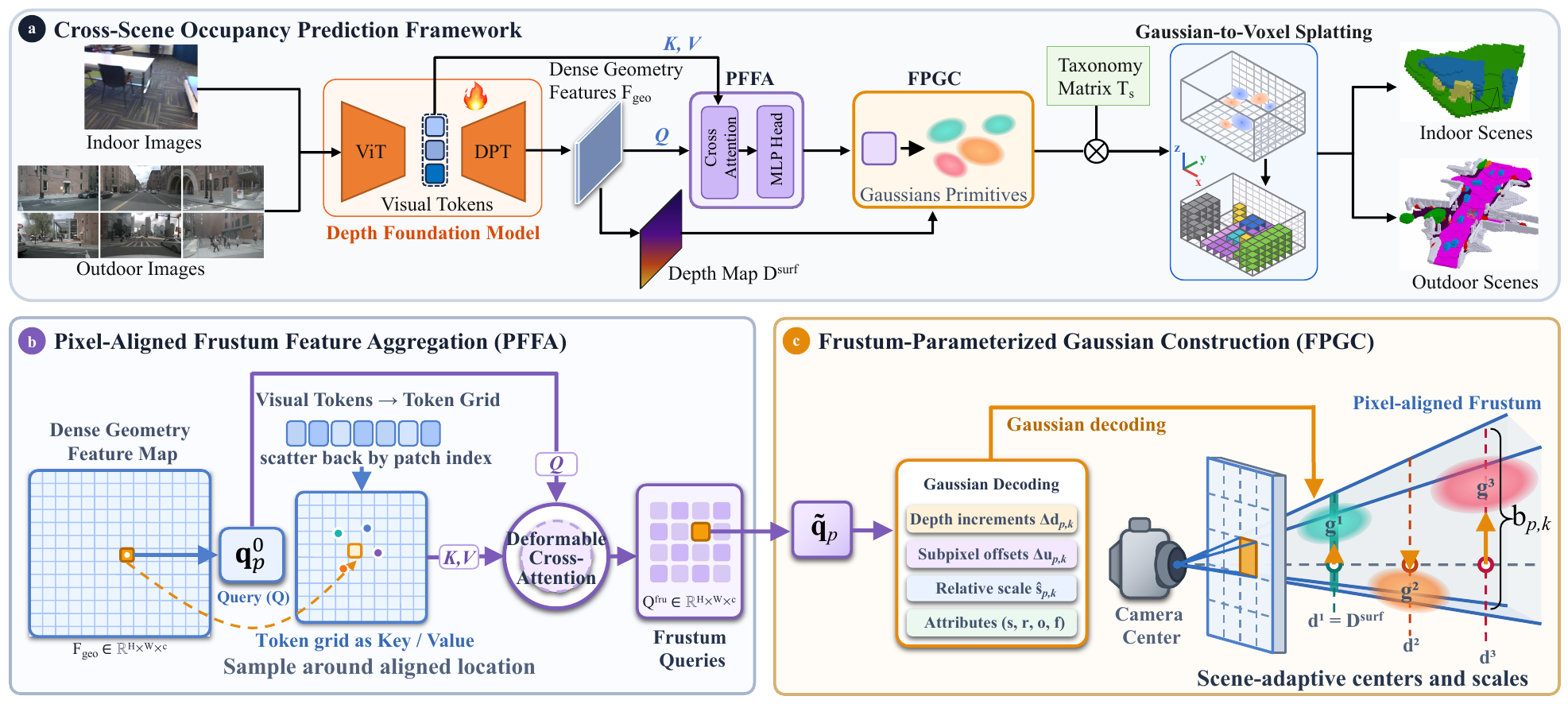}
\caption{Overview of OccAnyScene, a pixel-frustum-centered Gaussian framework. OccAnyScene treats each pixel frustum as the basic unit for constructing a continuous scene representation. \textbf{Pixel-Aligned Frustum Feature Aggregation} aggregates dense geometry features and visual tokens into frustum queries, while \textbf{Frustum-Parameterized Gaussian Construction} decodes each query into multiple Gaussians. The resulting Gaussians are adapted to scene-specific semantic taxonomies and splatted onto the target voxel grid to produce semantic occupancy predictions.}
\label{fig:main}
\end{figure*}

\subsection{Vision-Centric 3D Occupancy Prediction}
3D occupancy prediction aims to infer dense volumetric geometry and semantics from visual observations~\cite{hayes20253d,occsurvey}. Large-scale benchmarks such as Occ3D~\cite{tian2023occ3d} and OpenOccupancy~\cite{wang2023openoccupancy} have promoted vision-centric occupancy prediction in outdoor driving scenes based on nuScenes~\cite{caesar2020nuscenes} and Waymo~\cite{sun2020scalability}. Existing methods explore dense voxels~\cite{li2023voxformer,liu2024fully}, BEV/TPV representations~\cite{yu2023flashocc,huang2023tri}, 3D feature aggregation~\cite{yu2024context}, and rendering-based supervision~\cite{gan2025gaussianocc,gan2024comprehensive}. Indoor occupancy prediction has also been studied through monocular semantic scene completion~\cite{cao2022monoscene} and depth-guided indoor occupancy benchmarks~\cite{yu2024monocular}. Despite these advances, most methods are developed under scene-specific protocols, where the input setting, spatial range, voxel resolution, and semantic taxonomy are tied to a particular scene type or dataset. In contrast, OccAnyScene studies cross-scene occupancy prediction with one model across heterogeneous indoor and outdoor datasets.

\subsection{Gaussian-based Occupancy Prediction}
Dense voxel and BEV/TPV representations discretize predefined 3D domains, tying them to fixed spatial ranges and resolutions and making them restrictive for cross-scene settings with varying scene extents, voxel specifications, and camera geometry. Sparse Gaussian methods, including GaussianFormer~\cite{huang2024gaussianformer,huang2025gaussianformer}, EmbodiedOcc~\cite{wu2025embodiedocc}, and SplatSSC~\cite{qian2026splatssc}, offer continuous Gaussian primitives that can be splatted onto different target grids. However, their Gaussian construction remains scene-specific, relying on either predefined 3D ranges or depth-guided lifting. The former retains fixed global support, while the latter adapts Gaussian centers to scene geometry but concentrates them near visible surfaces and weakly constrains their scales, limiting stability across cameras and scene scales. These limitations motivate our pixel-frustum formulation, which provides a scene-adaptive geometric reference for Gaussian positions and scales across varying cameras and scene ranges while supporting coverage beyond visible surfaces.

\subsection{Multi-Source and Generalizable 3D Perception}
Recent 3D perception methods have begun to study multi-source learning and transferable scene understanding. Omni3D~\cite{brazil2023omni3d} explores multi-source training for 3D object detection and shows improved transfer to smaller target datasets. For occupancy prediction, OccAny~\cite{cao2026occany} studies generalized 3D occupancy in unconstrained urban scenes, but it relies on a reconstruction-rendering-fusion pipeline at test time. In contrast, OccAnyScene addresses Cross-Scene 3D Occupancy Prediction by directly predicting occupancy with a single feed-forward model, avoiding costly test-time reconstruction or fusion.

\section{Method}

\subsection{Cross-Scene Formulation and Overview}
\label{sec:overview}

In Cross-Scene 3D Semantic Occupancy Prediction, a single model is expected to predict semantic occupancy across heterogeneous scene types. Given monocular or surround-view RGB images with known camera intrinsics and extrinsics, the target occupancy space for a scene domain indexed by $s$ is defined by a 3D region $\Omega_s$, a voxel size $v_s$, and a scene-specific semantic taxonomy $\mathcal{C}_s$. The goal is to predict a semantic label from $\mathcal{C}_s$ for each voxel center in $\mathcal{Q}_{\Omega_S,v_S}=\{\mathbf{x}_m\}_{m=1}^{M}$, where $\mathbf{x}_m$ is the $m$-th voxel center and $M$ is the total number of voxels. To decouple scene representation from domain-specific output protocols, OccAnyScene represents each scene using continuous 3D Gaussians ~\cite{kerbl3Dgaussians}. The $i$-th Gaussian primitives are represented as $\mathbf{a}_i=[\boldsymbol{\mu}_i,\mathbf{s}_i,\mathbf{r}_i,o_i,\mathbf{f}_i]$, where $\boldsymbol{\mu}_i$, $\mathbf{s}_i$, $\mathbf{r}_i$, $o_i$, and $\mathbf{f}_i\in\mathbb{R}^{C_f}$ denote its center, scale, rotation, opacity logit, and shared semantic feature of dimension $C_f$, respectively.

As illustrated in Fig.~\ref{fig:main}, OccAnyScene treats the finite 3D frustum associated with each feature pixel as the basic unit for constructing the scene representation. From each input view, a pretrained depth foundation model ~\cite{yang2024depth, lin2025depth} provides visual tokens from its ViT encoder and a dense geometry feature map decoded from these tokens by its DPT head. \textbf{Pixel-Aligned Frustum Feature Aggregation (PFFA)} conditions each dense pixel feature and aggregates visual tokens through pixel-aligned deformable cross-attention, producing one frustum query per pixel. \textbf{Frustum-Parameterized Gaussian Construction (FPGC)} then directly decodes each frustum query into multiple final Gaussians. It uses surface-anchored depths to extend the representation coverage from visible surfaces into occluded regions and parameterizes the Gaussian center offsets and scales relative to the corresponding frustum geometry. Finally, Gaussians from all views are combined, their semantic features are mapped to $\mathcal{C}_s$, and they are splatted onto $\mathcal{Q}_{\Omega_s,v}$ to produce the occupancy prediction.

\subsection{Pixel-Aligned Frustum Feature Aggregation} 
\label{sec:pffa}
Representing a pixel frustum requires visible-surface geometry, image-to-ray projection geometry, and contextual cues for occlusion reasoning. For each input view, the pretrained depth foundation model provides the ViT encoder tokens and a dense geometry feature map \(\mathbf{F}_{\mathrm{geo}}\) decoded by its DPT head. Each location \(p\) in \(\mathbf{F}_{\mathrm{geo}}\) corresponds to an image patch and its associated 3D frustum: \(\mathbf{F}_{\mathrm{geo}}(p)\) provides pixel-aligned geometric cues, while the ViT tokens contain rich contextual information. We initialize the pixel query as \(\mathbf{q}_p^{0}
=
\phi_{\mathrm{geo}}
\left(
\mathbf{F}_{\mathrm{geo}}(p)
\right)
+
\phi_{\mathrm{cam}}
\left(
\mathbf{R}(p)
\right),\)
where \(\mathbf{R}(p)\in\mathbb{R}^{3}\) is the ray direction at feature pixel \(p\) derived from the camera intrinsics, and \(\phi_{\mathrm{geo}}\) and \(\phi_{\mathrm{cam}}\) are learnable projections.

To incorporate the local context required for occlusion reasoning, we restore the sequential ViT tokens to their two-dimensional token grids, collectively denoted by \(\mathbf{V}\), and perform pixel-aligned deformable cross-attention:
\[
\widetilde{\mathbf{q}}_p
=
\operatorname{DeformAttn}
\left(
\mathbf{q}_p^{0},
\boldsymbol{\rho}_p,
\mathbf{V}
\right),
\]
where \(\mathbf{q}_p^{0}\) serves as the query, \(\mathbf{V}\) as the keys and values, and \(\boldsymbol{\rho}_p\) is the normalized reference location of feature pixel \(p\). By sampling around \(\boldsymbol{\rho}_p\), the attention aggregates nearby semantic and structural cues for occlusion reasoning. The resulting \(\widetilde{\mathbf{q}}_p\) combines visible-surface geometry, camera-dependent viewing geometry, and local context, forming the frustum query subsequently decoded by FPGC.

\subsection{Frustum-Parameterized Gaussian Construction}
\label{sec:fpgc}
Given the camera-aware frustum query \(\widetilde{\mathbf{q}}_p\) produced by PFFA, Frustum-Parameterized Gaussian Construction (FPGC) predicts the intermediate parameters and attributes for constructing \(K\) Gaussians within the corresponding pixel frustum:
\[
\left\{
\Delta d_{p,k},
\Delta\mathbf{u}_{p,k},
\widehat{\mathbf{s}}_{p,k},
\mathbf{r}_{p,k},
o_{p,k},
\mathbf{f}_{p,k}
\right\}_{k=1}^{K}
=
\phi_{\mathrm{G}}
\left(
\widetilde{\mathbf{q}}_p
\right),
\]
where \(\phi_{\mathrm{G}}\) denotes an MLP-based Gaussian decoding head. For the \(k\)-th Gaussian, \(\Delta d_{p,k}\), \(\Delta\mathbf{u}_{p,k}\), and \(\widehat{\mathbf{s}}_{p,k}\) denote its surface-relative depth increment, subpixel offset, and relative scale, respectively. Specifically, softplus enforces \(\Delta d_{p,k}>0\) and \(\widehat{\mathbf{s}}_{p,k}\in\mathbb{R}_{>0}^{3}\), while tanh bounds \(\Delta\mathbf{u}_{p,k}\) to \([-1,1]^2\) in local feature-pixel coordinates. The remaining outputs \(\mathbf{r}_{p,k}\), \(o_{p,k}\), and \(\mathbf{f}_{p,k}\) denote its rotation, opacity logit, and semantic feature, respectively.

\noindent\textbf{Canonical-camera depth prediction.} To provide a common visible-surface anchor for these Gaussians, we predict a surface depth map from the dense geometry feature map. Since varying camera intrinsics can cause metric inconsistency across datasets, we follow Metric3Dv2~\cite{hu2024metric3d} and perform depth prediction in a canonical camera space:
\[
\mathbf{D}^{\mathrm{can}}
=
\phi_{\mathrm{depth}}
\left(
\mathbf{F}_{\mathrm{geo}}
\right),
\qquad
\mathbf{D}^{\mathrm{surf}}
=
\mathbf{D}^{\mathrm{can}}
\frac{f_{\mathrm{real}}}{f_{\mathrm{can}}},
\]
where \(\phi_{\mathrm{depth}}\) is a lightweight MLP head, and \(f_{\mathrm{real}}\) and \(f_{\mathrm{can}}\) denote the effective focal lengths of the input and canonical cameras, respectively. The resulting metric surface depth \(d_p^{\mathrm{surf}}=\mathbf{D}^{\mathrm{surf}}(p)\) is used as the common depth anchor for the \(K\) Gaussians associated with feature pixel \(p\).

\begin{figure*}[t]
\centering
\includegraphics[width=\textwidth]{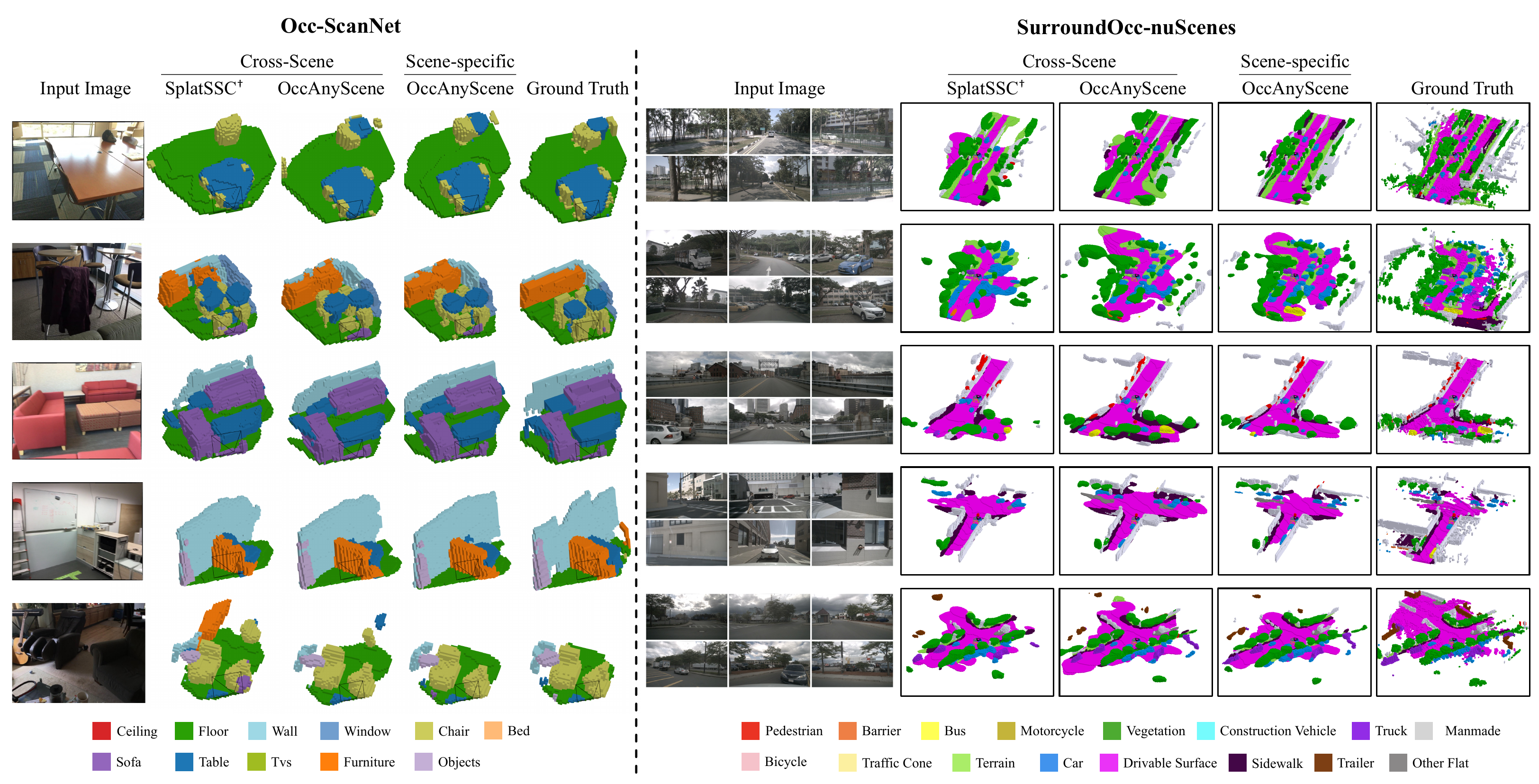}
\caption{Qualitative results on Occ-ScanNet and SurroundOcc-nuScenes. We compare scene-specific and cross-scene OccAnyScene-DAv2 with SplatSSC\({}^{\dagger}\). \({}^{\dagger}\) denotes our cross-scene adaptation of SplatSSC.}
\label{fig:fig_vis}
\end{figure*}

\noindent\textbf{Pixel-Frustum-Constrained Gaussian Positioning.} Although the predicted surface depth provides a geometry-adaptive anchor, placing all Gaussians at this depth would restrict the representation to visible surfaces. We therefore predict a separate depth increment \(\Delta d_{p,k}\) for each Gaussian relative to the common surface anchor, allowing the \(K\) Gaussians to extend along the frustum into occluded regions. Meanwhile, each discrete feature pixel represents a finite image-plane region and therefore corresponds to a 3D frustum rather than only its central ray. As \(\mathbf{F}_{\mathrm{geo}}\) is spatially downsampled, each feature pixel covers a relatively coarse patch of the input image. We thus predict a bounded subpixel offset \(\Delta\mathbf{u}_{p,k},\) for each Gaussian, allowing its center to move laterally within the corresponding pixel frustum. Accordingly, the depth and image-plane location of the \(k\)-th Gaussian are given by
\[
d_{p,k}
=
d_p^{\mathrm{surf}}
+
\Delta d_{p,k},
\qquad
\mathbf{u}_{p,k}
=
\mathbf{u}_p
+
\Delta\mathbf{u}_{p,k},
\]
where \(\mathbf{u}_p\) is the center of feature pixel \(p\). Different depth increments \(\Delta d_{p,k}\) allow the \(K\) Gaussians to extend from their common surface anchor into occluded regions, while the bounded offsets \(\Delta\mathbf{u}_{p,k}\) refine the coarse pixel-center localization before back-projection.

The 3D center of each Gaussian is obtained by back-projecting its image-plane location and depth:
\[
\boldsymbol{\mu}_{p,k}
=
\Pi^{-1}
\left(
\mathbf{u}_{p,k},
d_{p,k};
\mathbf{P}
\right),
\]
where \(\Pi^{-1}\) denotes the inverse projection of the camera, and \(\mathbf{P}\) is the projection matrix that includes intrinsics and extrinsics of the camera.
 
\noindent\textbf{Pixel-Frustum-Relative Scale Parameterization.} Directly regressing Gaussian scales in absolute metric units is unsuitable for cross-scene learning, because their magnitudes vary substantially with camera intrinsics, scene depths, and spatial ranges. Instead, we use the metric cross-section of the pixel frustum at each Gaussian depth as an adaptive scale reference:
\[
b_{p,k}
=
\eta
\cdot
\frac{1}{2}
\left(
\frac{d_{p,k}}{f_x}
+
\frac{d_{p,k}}{f_y}
\right),
\qquad
\mathbf{s}_{p,k}
=
b_{p,k}
\widehat{\mathbf{s}}_{p,k},
\]
where \(f_x\) and \(f_y\) are the focal lengths rescaled to the resolution of \(\mathbf{F}_{\mathrm{geo}}\), \(b_{p,k}\) is the scalar frustum scale reference, and \(\eta\) is a fixed hyperparameter controlling its base magnitude. The decoder predicts the dimensionless relative scale \(\widehat{\mathbf{s}}_{p,k}\in\mathbb{R}^{3}\), which determines the three-axis Gaussian scale relative to \(b_{p,k}\). This parameterization adapts Gaussian spatial supports to different camera parameters and scene ranges without directly regressing cross-scene absolute scales.

With the above parameterization, the \(k\)-th final Gaussian constructed from feature pixel \(p\) is represented as
\[
\mathbf{a}_{p,k}
=
\left[
\boldsymbol{\mu}_{p,k},
\mathbf{s}_{p,k},
\mathbf{r}_{p,k},
o_{p,k},
\mathbf{f}_{p,k}
\right].
\]

\begin{table}[t]
\centering
\scriptsize
\setlength{\tabcolsep}{2.5pt}
\vspace{-2mm}
\resizebox{\columnwidth}{!}{%
\begin{tabular}{l|c|c|c|c|c} 
\toprule
Dataset & Scene & Input & Samples & Range & Vox. \\
\midrule
Occ-ScanNet & Indoor & Mono & 47.5K & $4.8{\times}4.8{\times}2.88$ m & $0.08$ m \\
SurroundOcc & Outdoor & Surround & 34.1K & $100{\times}100{\times}8$ m & $0.5$ m \\
\bottomrule
\end{tabular}
}
\caption{
Comparison of the datasets used in our experiments.
}
\label{tab:dataset_comparison}
\vspace{-4mm}
\end{table}

\subsection{Gaussian-to-Occ and Training Objectives}
\label{sec:taxonomy_g2v}

The final Gaussians constructed by FPGC from all feature pixels and input views form a shared continuous scene representation. To accommodate different semantic taxonomies, the shared semantic feature \(\mathbf{f}_{p,k}\) of each Gaussian is mapped to the label space of scene domain \(s\) using a learnable taxonomy matrix \(\mathbf{T}_s\in\mathbb{R}^{C_s\times C_f}\), where \(C_s\) is the number of semantic classes. Specifically, the domain-specific semantic logits of \(\mathbf{a}_{p,k}\) are computed as \(\mathbf{T}_s\mathbf{f}_{p,k}\in\mathbb{R}^{C_s}\). This formulation maintains a shared Gaussian semantic space while adapting its predictions to different taxonomies.

Given the constructed Gaussians and their scene-specific semantic logits, we convert the continuous representation into the target voxel grid using the Decoupled Gaussian Aggregator (DGA) from SplatSSC~\cite{qian2026splatssc}. Following SplatSSC~\cite{qian2026splatssc}, we supervise the predicted occupancy using focal loss, Lovasz-Softmax loss, and Probability Scale Loss:
\[
\mathcal{L}_{\mathrm{occ}}
=
\lambda_{\mathrm{focal}}\mathcal{L}_{\mathrm{focal}}
+
\lambda_{\mathrm{lov}}\mathcal{L}_{\mathrm{lov}}
+
\lambda_{\mathrm{scal}}\mathcal{L}_{\mathrm{scal}}^{\mathrm{prob}}.
\]

\begin{table*}[t] %
        \vspace{-4mm}
		\small
            \setlength{\tabcolsep}{0.004\linewidth}  
            \begin{center}
           \resizebox{1.0\linewidth}{!}{
		\begin{tabular}{l|c|c|c|c c c c c c c c c c c}
			\toprule
			Method
			& Setting
			& IoU
			& {mIoU}
			& \rotatebox{90}{\parbox{1.5cm}{\textcolor{ceiling}{$\blacksquare$} ceiling}} 
			& \rotatebox{90}{\textcolor{floor}{$\blacksquare$} floor}
			& \rotatebox{90}{\textcolor{wall}{$\blacksquare$} wall} 
			& \rotatebox{90}{\textcolor{window}{$\blacksquare$} window} 
			& \rotatebox{90}{\textcolor{chair}{$\blacksquare$} chair} 
			& \rotatebox{90}{\textcolor{bed}{$\blacksquare$} bed} 
			& \rotatebox{90}{\textcolor{sofa}{$\blacksquare$} sofa} 
			& \rotatebox{90}{\textcolor{table}{$\blacksquare$} table} 
			& \rotatebox{90}{\textcolor{tvs}{$\blacksquare$} tvs} 
			& \rotatebox{90}{\textcolor{furniture}{$\blacksquare$} furniture} 
			& \rotatebox{90}{\textcolor{objects}{$\blacksquare$} objects}\\
			\midrule
			MonoScene (\citeyear{cao2022monoscene}) & Scene-specific & 41.60 & 24.62 & 15.17 & 44.71 & 22.41 & 12.55 & 26.11 & 27.03 & 35.91 & 28.32 & 6.57 & 32.16 & 19.84 \\
            ISO (\citeyear{yu2024monocular}) & Scene-specific & 42.16 & 28.71 & 19.88 & 41.88 & 22.37 & 16.98 & 29.09 & 42.43 & 42.00 & 29.60 & 10.62 & 36.36 & 24.61 \\
            EmbodiedOcc (\citeyear{wu2025embodiedocc}) & Scene-specific & 53.95 & 42.90 & 40.90 & 50.80 & 41.90 & 33.00 & 41.20 & 55.20 & 61.90 & 43.80 & 35.40 & 53.50 & 42.90 \\
            EmbodiedOcc++ (\citeyear{wang2025embodiedocc++}) & Scene-specific & 54.90 & 46.20 & 36.40 & 53.10 &  41.80 & 34.40 & 42.90 & 57.30 & 64.10 & 45.20 & 34.80 & 54.20 & 44.10 \\
            RoboOcc (\citeyear{zhang2025roboocc}) & Scene-specific & 56.48 & 47.67 & 45.36 & 53.49 & 44.35 & 34.81 & 43.38 & 56.93 & 63.35 & 46.35 & 36.12 & 55.48 & 44.78 \\
            GPOcc (\citeyear{zhou2026generalizing}) & Scene-specific & 63.14 & 56.19 & 51.67 &  59.93 & 52.07 & 46.44 & 51.35 & 64.45 & 69.47 & 54.30 & 51.76 & 63.29 & 53.36 \\
            SplatSSC (\citeyear{qian2026splatssc}) & Scene-specific & 62.83 & 51.83 & 49.10 & 59.00 & 48.30 & 38.80 & 47.40 & 62.40 & 67.00 & 49.50 & 42.60 & 60.70 & 45.40 \\
            SplatSSC$^\dagger$ (\citeyear{qian2026splatssc}) & Cross-scene & 57.42 & 46.80 & 41.1 & 54.5 & 40.10 & 32.30 & 42.20 & 59.90 & 64.60 & 44.60 & 37.60 & 56.30 & 41.50 \\
            \midrule            
            OccAnyScene-DAv2 & Scene-specific & 64.98 & 55.42 & 52.40 & 62.20 & 47.40 & 41.40 & 53.00 & 62.20 & 72.70 & 55.90 & 41.10 & 64.40 & 51.90 \\
            OccAnyScene-DAv2 & Cross-scene & 64.56 & 55.46 & 52.30 & 61.30 & 47.90 & 42.50 & 51.30 & 66.40 & 71.80 & 54.40 & 47.60 & 63.70 & 50.90 \\

            \midrule
            OccAnyScene-DAv3 & Scene-specific & \textbf{68.34}  & \textbf{59.92} & \underline{57.50} & \textbf{65.40} & \textbf{52.70} & \textbf{47.90} & \textbf{55.10} & \textbf{70.50} & \textbf{74.90} & \textbf{59.10} & \textbf{52.60} & \textbf{67.60} & \textbf{55.70} \\
 
            OccAnyScene-DAv3 & Cross-scene & \underline{67.96} & \underline{59.51} & \textbf{59.10} & \underline{64.40} & \underline{52.50} & \underline{47.10} & \underline{54.90} & \underline{70.00} & \underline{74.50} & \underline{58.10} & \underline{51.60} & \underline{67.20} & \underline{55.10} \\            
			\bottomrule
		\end{tabular}
		}
            \end{center}
            \vspace{-.5em}
            \caption{
                    Comparison with state-of-the-art methods on Occ-ScanNet for 3D semantic occupancy prediction. ``Scene-specific'' denotes training only on Occ-ScanNet, while ``Cross-scene'' denotes one shared model jointly trained on Occ-ScanNet and SurroundOcc-nuScenes. \({}^\dagger\) denotes our cross-scene adaptation of SplatSSC with necessary dataset-specific output adapters, while no proposed OccAnyScene modules are used. The best and second-best results are highlighted in \textbf{bold} and \underline{underlined}.
                }
            \captionsetup{font=scriptsize}
            \label{tab:mono_scannet}
\end{table*}

We additionally supervise the predicted surface depth \(\mathbf{D}_{\mathrm{surf}}\) with a robust Huber loss \(\mathcal{L}_{\mathrm{depth}}\). Unlike the two-stage training adopted by SplatSSC, OccAnyScene jointly optimizes depth estimation and occupancy prediction with a single end-to-end objective:
\[
\mathcal{L}
=
\mathcal{L}_{\mathrm{occ}}
+
\lambda_{\mathrm{depth}}\mathcal{L}_{\mathrm{depth}}.
\]

\section{Experiments}

\suppressfloats[t]

\subsection{Datasets and Evaluation Metrics}
We conduct experiments on Occ-ScanNet~\cite{yu2024monocular}, SurroundOcc-nuScenes~\cite{wei2023surroundocc}, covering both indoor and outdoor scenes with diverse camera configurations, spatial ranges, and voxel resolutions, as summarized in Tab.~\ref{tab:dataset_comparison}. These datasets enable evaluation under both single-dataset and multi-source joint training settings. Following standard occupancy evaluation protocols, we report occupancy IoU for geometric occupancy prediction and semantic mIoU for semantic occupancy prediction.

\begin{figure}[t]
\centering
\includegraphics[width=\linewidth]{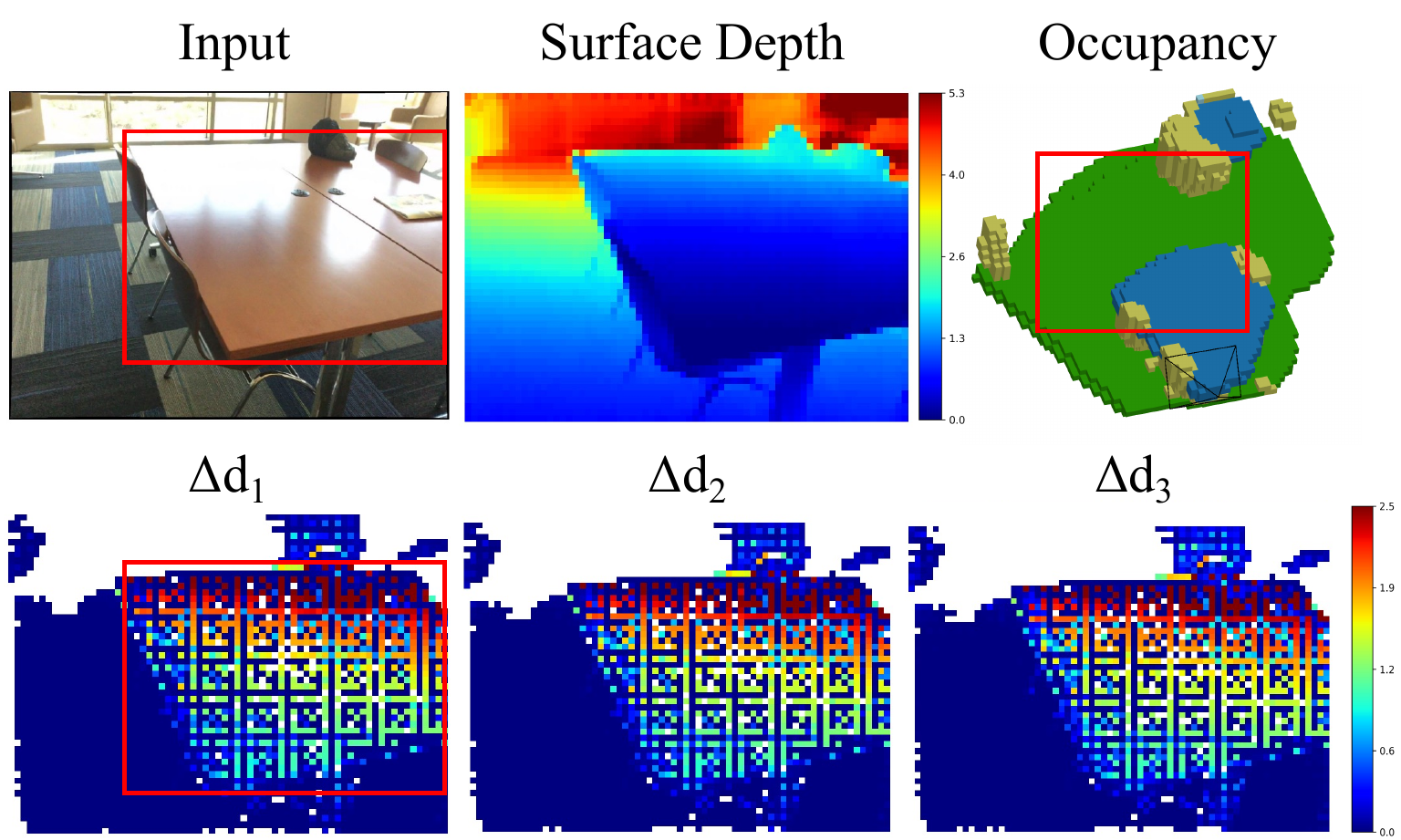}
\caption{Per-Gaussian depth increments on Occ-ScanNet. With \(K=3\), FPGC constructs three Gaussians within each pixel frustum, each with a predicted depth increment relative to the predicted surface depth. The bottom row visualizes these increments after per-pixel sorting. Red boxes highlight foreground-occlusion regions.}
\label{fig:fig_layer_depth}
\vspace{-4mm}
\end{figure}

\subsection{Implementation Details}

We use DepthAnythingV2 (DAv2)~\cite{yang2024depth} with ViT-Base as the default image encoder for feature extraction, and also evaluate a stronger DepthAnythingV3 (DAv3)~\cite{lin2025depth} variant with ViT-Large. To reduce the computational cost of cross-attention, we downsample the dense geometry feature map \(\mathbf{F}_{\mathrm{geo}}\) from the depth foundation model to \(1/8\) of the input resolution and set \(K=3\) Gaussians per pixel frustum by default. Occ-ScanNet~\cite{yu2024monocular} uses monocular RGB images as input, while SurroundOcc-nuScenes~\cite{wei2023surroundocc} uses six surround-view camera images. For multi-source joint training, we alternate among datasets and keep the effective number of iterations for each dataset identical to its single-dataset setting, ensuring a fair comparison with scene-specific models. In our cross-scene model, only the taxonomy matrices are dataset-specific, while all remaining parameters are shared between Occ-ScanNet and SurroundOcc-nuScenes. To accommodate their different semantic spaces, each Gaussian predicts a shared \(32\)-dimensional semantic feature, which is mapped to dataset-specific semantic logits by the corresponding taxonomy matrix.

\begin{table*}[t] %
    \small
    \setlength{\tabcolsep}{0.004\linewidth}  
    \renewcommand\arraystretch{1.2}
    \centering
    \vspace{-.5em}
    \resizebox{\textwidth}{!}{
    \begin{tabular}{l|c|c c | c c c c c c c c c c c c c c c c}
        \toprule
        Method
        & Setting
        & IoU
        & mIoU
        & \rotatebox{90}{\textcolor{nbarrier}{$\blacksquare$} barrier}
        & \rotatebox{90}{\textcolor{nbicycle}{$\blacksquare$} bicycle}
        & \rotatebox{90}{\textcolor{nbus}{$\blacksquare$} bus}
        & \rotatebox{90}{\textcolor{ncar}{$\blacksquare$} car}
        & \rotatebox{90}{\textcolor{nconstruct}{$\blacksquare$} const. veh.}
        & \rotatebox{90}{\textcolor{nmotor}{$\blacksquare$} motorcycle}
        & \rotatebox{90}{\textcolor{npedestrian}{$\blacksquare$} pedestrian}
        & \rotatebox{90}{\textcolor{ntraffic}{$\blacksquare$} traffic cone}
        & \rotatebox{90}{\textcolor{ntrailer}{$\blacksquare$} trailer}
        & \rotatebox{90}{\textcolor{ntruck}{$\blacksquare$} truck}
        & \rotatebox{90}{\textcolor{ndriveable}{$\blacksquare$} drive. suf.}
        & \rotatebox{90}{\textcolor{nother}{$\blacksquare$} other flat}
        & \rotatebox{90}{\textcolor{nsidewalk}{$\blacksquare$} sidewalk}
        & \rotatebox{90}{\textcolor{nterrain}{$\blacksquare$} terrain}
        & \rotatebox{90}{\textcolor{nmanmade}{$\blacksquare$} manmade}
        & \rotatebox{90}{\textcolor{nvegetation}{$\blacksquare$} vegetation}
        \\
        \midrule


        TPVFormer (\citeyear{huang2023tri}) & Scene-specific & {30.86} & 17.10 & 15.96 & 5.31 & 23.86 & 27.32 & 9.79 & 8.74 & 7.09 & 5.20 & 10.97 & 19.22 & {38.87} & {21.25} & {24.26} & {23.15} & 11.73 & 20.81 \\

        OccFormer (\citeyear{zhang2023occformer}) & Scene-specific & {31.39} & {19.03} & {18.65} & {10.41} & {23.92} & {30.29} & {10.31} & {14.19} & {13.59} & {10.13} & {12.49} & {20.77} & {38.78} & 19.79 & 24.19 & 22.21 & {13.48} & {21.35} \\
        
        SurroundOcc (\citeyear{wei2023surroundocc}) & Scene-specific & {31.49} & {20.30} & {20.59} & {11.68} & {28.06} & {30.86} & {10.70} & {15.14} & {14.09} & {12.06} & {14.38} & {22.26} & 37.29 & {23.70} & {24.49} & {22.77} & {14.89} & {21.86} \\

        GaussianFormer (\citeyear{huang2024gaussianformer}) & Scene-specific & 29.83 & {19.10} & {19.52} & {11.26} & {26.11} & {29.78} & {10.47} & {13.83} & {12.58} & {8.67} & {12.74} & {21.57} & {39.63} & {23.28} & {24.46} & {22.99} & 9.59 & 19.12 \\
        
        GaussianFormer-2 (\citeyear{huang2025gaussianformer}) & Scene-specific & 31.74 & 20.82 & 21.39 & 13.44 & 28.49 & 30.82 & 10.92 & 15.84 & 13.55 & 10.53 & 14.04 & 22.92 & 40.61 & 24.36 & 26.08 & 24.27 & 13.83 & 21.98 \\

        GaussianWorld (\citeyear{zuo2025gaussianworld}) & Scene-specific & 32.77 & 21.79 & \underline{21.61} & 13.30 & 27.28 & \underline{31.21} & 13.89 & 16.91 & 13.28 & 11.77 & 14.82 & 23.66 & 41.91 & 24.31 & \textbf{28.35} & 26.32 & 15.67 & 24.54 \\

        VG3S (\citeyear{yan2026vg3s}) & Scene-specific & 34.41 & 21.52 & 20.78 & 12.40 & 28.9 & 29.53 & 11.93 & 15.59 & 12.38 & 10.61 & 14.65 & 21.74 & \underline{42.42} & 26.39 & \underline{28.06} & 26.58 & 17.46 & 25.76 \\

        DLWM (\citeyear{zhu2026dlwm}) & Scene-specific & 34.61 & 21.85 & \textbf{21.71} & 13.56 & 27.65 & \textbf{31.22} & 12.50 & 16.93 & 12.35 & 11.42 & 13.96 & 24.12 & \textbf{42.27} & 24.94 & 27.39 & 26.36 & 16.61 & 26.64 \\
        
        SplatSSC$^\dagger$ (\citeyear{qian2026splatssc}) & Scene-specific & 32.19 & 19.05 & 16.20 & 11.10 & 25.50 & 25.50 & 12.20 & 13.60 & 11.40 & 8.50 & 12.80 & 19.20 & 37.30 & 22.50 & 25.20 & 24.80 & 15.30 & 23.80 \\
        
        SplatSSC$^\dagger$ (\citeyear{qian2026splatssc}) & Cross-scene & 31.24 & 17.86 & 14.40 & 10.20 & 24.30 & 24.00 & 12.10 & 12.50 & 10.30 & 7.40 & 11.90 & 18.60 & 37.2 & 19.30 & 23.80 & 23.30 & 14.00 & 22.80 \\
   
        \midrule
        OccAnyScene-DAv2 & Scene-specific & 33.86 & 20.46 & 17.90 & 12.60 & 27.20 & 27.70 & 12.70 & 15.30 & 12.80 & 10.40 & 13.80 & 20.90 & 39.50 & 24.10 & 25.60 & 25.10 & 16.80 & 25.20 \\

        OccAnyScene-DAv2 & Cross-scene & 33.76 & 20.36 & 18.00 & 13.00 & 26.70 & 27.50 & 12.20 & 14.80 & 13.10 & 10.50 & 13.50 & 20.20 & 39.70 & 23.40 & 25.90 & 25.40 & 16.70 & 25.10 \\

        \midrule
        OccAnyScene-DAv3 & Scene-specific & \textbf{35.97} & \textbf{23.06} & 21.30 & \textbf{15.60} & \underline{29.90} & 30.70 & \textbf{14.90} & \textbf{17.60} & \underline{14.70} & \textbf{12.30} & \textbf{16.90} & \textbf{24.30} & 42.20 & \textbf{27.10} & 27.80 & \textbf{26.80} & \underline{19.10} & \textbf{27.60} \\
        
        OccAnyScene-DAv3 & Cross-scene & \underline{35.83} & \underline{22.87}  & 21.40 & \underline{14.80} & \textbf{30.00} & 30.20 & \underline{14.90} & \underline{17.00} & \textbf{14.80} & \underline{12.20} & \underline{16.80} & \underline{24.30} & 42.10 & \underline{26.60} & 27.50 & \underline{26.70} & \textbf{19.30} & \underline{27.50} \\
        
        \bottomrule
    \end{tabular}}
    \caption{
    Comparison with state-of-the-art methods on the SurroundOcc-nuScenes validation set. ``Scene-specific'' denotes training only on SurroundOcc-nuScenes, while ``Cross-scene'' denotes one shared model jointly trained on Occ-ScanNet and SurroundOcc-nuScenes. \({}^\dagger\) denotes our adapted SplatSSC-style baseline with necessary dataset-specific output adapters for SurroundOcc-nuScenes and cross-scene training, while no proposed OccAnyScene modules are used. The best and second-best results are highlighted in \textbf{bold} and \underline{underlined}, respectively.
    }
    \label{tab:nuscenes-results}
\end{table*}

\subsection{Main Results}

Tables~\ref{tab:mono_scannet} and~\ref{tab:nuscenes-results} compare scene-specific models trained separately on each benchmark with cross-scene models jointly trained on both datasets. OccAnyScene achieves strong performance under both settings, whereas directly adapting a scene-specific Gaussian occupancy model to heterogeneous indoor--outdoor training causes clear degradation. Specifically, the cross-scene variant of SplatSSC~\cite{qian2026splatssc} exhibits absolute mIoU drops of approximately \(5.03\%\) on Occ-ScanNet and \(1.19\%\) on SurroundOcc-nuScenes compared with its scene-specific counterparts. In contrast, OccAnyScene remains stable across both DAv2~\cite{yang2024depth} and DAv3~\cite{lin2025depth} encoders. With DAv3, joint cross-scene training reduces mIoU by only \(0.41\%\) on Occ-ScanNet and \(0.19\%\) on SurroundOcc-nuScenes. These results demonstrate that OccAnyScene can consolidate room-scale and street-scale occupancy prediction into a single model with negligible performance degradation.

\begin{table}[t]
\centering
\small
\vspace{-2mm}
\begin{tabular}{cc|cc|cc}
\toprule
PFFA & FPGC
& \multicolumn{2}{c|}{Occ-ScanNet}
& \multicolumn{2}{c}{nuScenes} \\
& & IoU & mIoU & IoU & mIoU \\
\midrule
           &            & 56.43 & 47.21 & 29.67 & 17.39 \\
\checkmark &            & 57.51 & 47.46 & 30.34 & 17.77 \\
           & \checkmark & 61.11 & 51.18 & 33.09 & 19.60 \\
\checkmark & \checkmark & \textbf{64.56} & \textbf{55.46} & \textbf{33.76} & \textbf{20.36} \\
\bottomrule
\end{tabular}
\caption{Ablation study of PFFA and FPGC under joint cross-scene training on Occ-ScanNet and SurroundOcc-nuScenes.}
\label{tab:ablation_module}

\end{table}

\begin{table}[t]
\centering
\small
\resizebox{\columnwidth}{!}{
\begin{tabular}{l|cc|cc}
\toprule
\multirow{2}{*}{Variant}
& \multicolumn{2}{c|}{Occ-ScanNet}
& \multicolumn{2}{c}{nuScenes} \\
& IoU & mIoU
& IoU & mIoU \\
\midrule
w/o canonical-camera depth & 62.61 & 53.37 & 32.85 & 19.30 \\
w/o depth residual              & 58.65 & 49.54 & 31.31 & 18.26 \\
w/o subpixel offset             & 63.68 & 54.55 & 33.14 & 19.91 \\
w/o frustum-relative scale      & 61.53 & 51.70 & 33.03 & 19.52 \\
\midrule
Full model                      & 64.56 & 55.46 & 33.76 & 20.36 \\
\bottomrule
\end{tabular}
}
\caption{Fine-grained ablation of Frustum-Parameterized Gaussian Construction under joint cross-scene training. Each variant removes one design from the full model.}
\vspace{-2mm}
\label{tab:ablation_fpgc}

\end{table}

\begin{table}[t]
\centering
\small
\vspace{-2mm}
\begin{tabular}{c|cc|cc}
\toprule
\multirow{2}{*}{\(K\)}
& \multicolumn{2}{c|}{Occ-ScanNet}
& \multicolumn{2}{c}{nuScenes} \\
& IoU & mIoU & IoU & mIoU \\
\midrule
1 & 64.37 & 54.80 & 33.36 & 20.30 \\
2 & 64.42 & 55.32 & 33.37 & 20.55 \\
3 & 64.56 & 55.46 & 33.76 & 20.36 \\
\bottomrule
\end{tabular}
\caption{Effect of the number \(K\) of Gaussians decoded from each pixel frustum under joint cross-scene training.}
\vspace{-2mm}
\label{tab:ablation_gaussian_number}

\end{table}

\subsection{Ablation Studies}

\noindent\textbf{Component ablation.} We conduct the ablation under joint cross-scene training using DAv2 as the image encoder. The baseline replaces DAv2's original per-pixel depth prediction head with a Gaussian prediction head that directly decodes each feature in \(\mathbf{F}_{\mathrm{geo}}\) into one Gaussian, without PFFA or FPGC. As shown in Table~\ref{tab:ablation_module}, PFFA alone brings limited improvement, whereas FPGC increases IoU by \(4.68\%\) on Occ-ScanNet and \(3.42\%\) on nuScenes. Combining PFFA with FPGC further enlarges the gains to \(8.13\%\) and \(4.09\%\), respectively, showing that PFFA is more effectively utilized when coupled with FPGC. This synergy highlights the importance of its frustum-based parameterization of Gaussian positions and scales.

\noindent\textbf{FPGC design ablation.} Table~\ref{tab:ablation_fpgc} evaluates the individual designs of FPGC. Removing any component consistently degrades performance on both benchmarks. Without depth increments, IoU decreases by \(5.91\%\) on Occ-ScanNet and \(2.45\%\) on nuScenes, confirming their importance for extending Gaussians beyond visible surfaces. Removing the frustum-relative scale parameterization causes IoU drops of \(3.03\%\) and \(0.73\%\), respectively, while canonical-camera depth and subpixel offsets also provide consistent improvements. The full model performs best across all metrics, demonstrating that these designs jointly enable stable Gaussian construction across heterogeneous scenes.

\begin{figure}[!t]
\centering
\includegraphics[width=\linewidth]{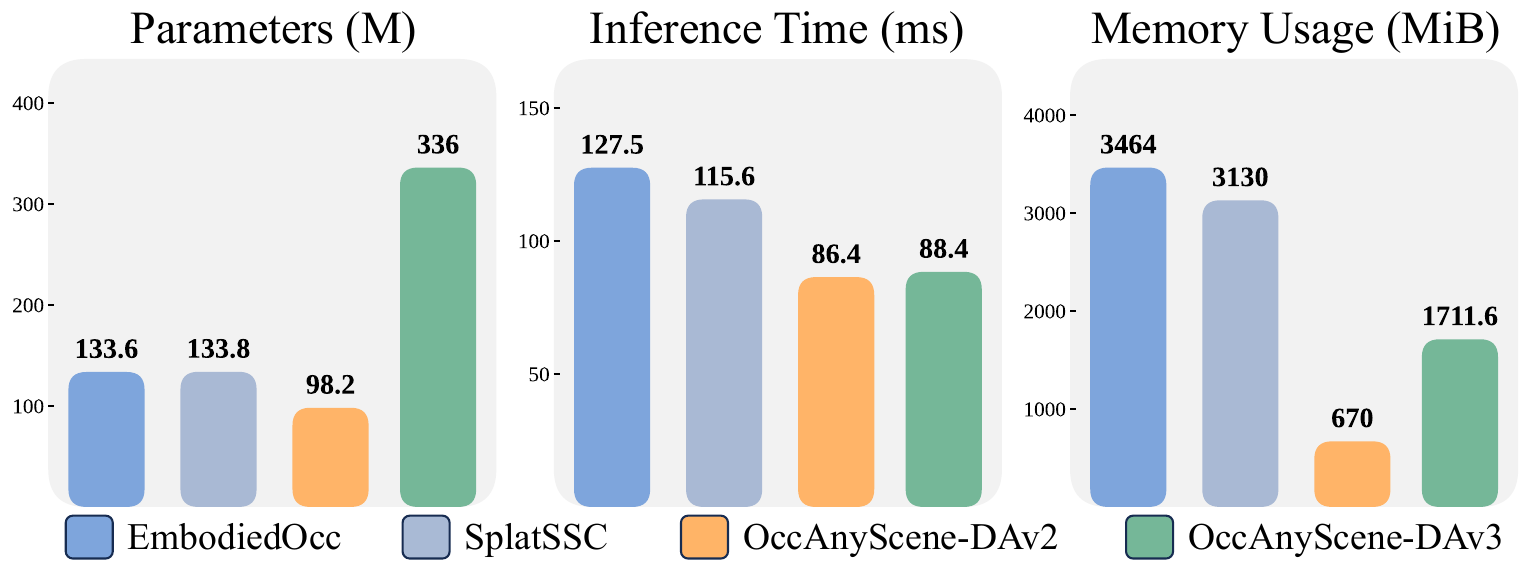}
\caption{Efficiency comparison on Occ-ScanNet in terms of parameter count, inference time, and memory usage. All measurements are obtained on a single NVIDIA RTX 4090 GPU. EmbodiedOcc and SplatSSC results are taken from SplatSSC~\cite{qian2026splatssc}.}
\label{fig:efficiency}
\end{figure}

\noindent\textbf{Number of Gaussians per pixel frustum.} Table~\ref{tab:ablation_gaussian_number} studies the effect of decoding \(K\) Gaussians from each pixel frustum. Increasing \(K\) from 1 to 3 brings only marginal gains of \(0.66\%\) mIoU on Occ-ScanNet and \(0.40\%\) IoU on nuScenes. To explain this limited sensitivity, Fig.~\ref{fig:fig_layer_depth} visualizes the surface-relative depth increments of the \(K=3\) Gaussians decoded from each pixel frustum. The three maps are highly similar, indicating that Gaussians decoded from the same frustum remain close rather than form well-separated depth layers. Meanwhile, foreground regions exhibit alternating increments across neighboring pixels: some pixels preserve their Gaussians near the visible surface, while others shift them toward the occluded background. The completion of the excluded space is therefore collectively achieved by Gaussians decoded from neighboring pixels, explaining why \(K=1\) remains competitive.

\subsection{Efficiency analysis}
Fig.~\ref*{fig:efficiency} compares OccAnyScene with EmbodiedOcc~\cite{wu2025embodiedocc} and SplatSSC~\cite{qian2026splatssc} on Occ-ScanNet. OccAnyScene-DAv2 achieves the best overall efficiency with \(98.2\)M parameters, \(86.4\) ms inference time, and \(670\) MiB memory, reducing memory usage by \(80.7\%\) and \(78.6\%\) relative to EmbodiedOcc and SplatSSC, respectively. Although DAv3 substantially increases model capacity, it retains a comparable inference time of \(88.4\) ms, providing a flexible accuracy--efficiency trade-off.

\subsection{Qualitative results}
Fig.~\ref{fig:fig_vis} compares the occupancy predictions of scene-specific OccAnyScene, cross-scene SplatSSC\({}^{\dagger}\), and cross-scene OccAnyScene on Occ-ScanNet and SurroundOcc-nuScenes. The single cross-scene OccAnyScene model remains qualitatively comparable to its separately trained scene-specific counterparts while more closely matching the ground truth than SplatSSC\({}^{\dagger}\) in both indoor structures and outdoor layouts. These observations are consistent with the quantitative results in Tabs.~\ref{tab:mono_scannet} and~\ref{tab:nuscenes-results}.

\begin{figure*}[!t]
\centering
\includegraphics[width=\textwidth]{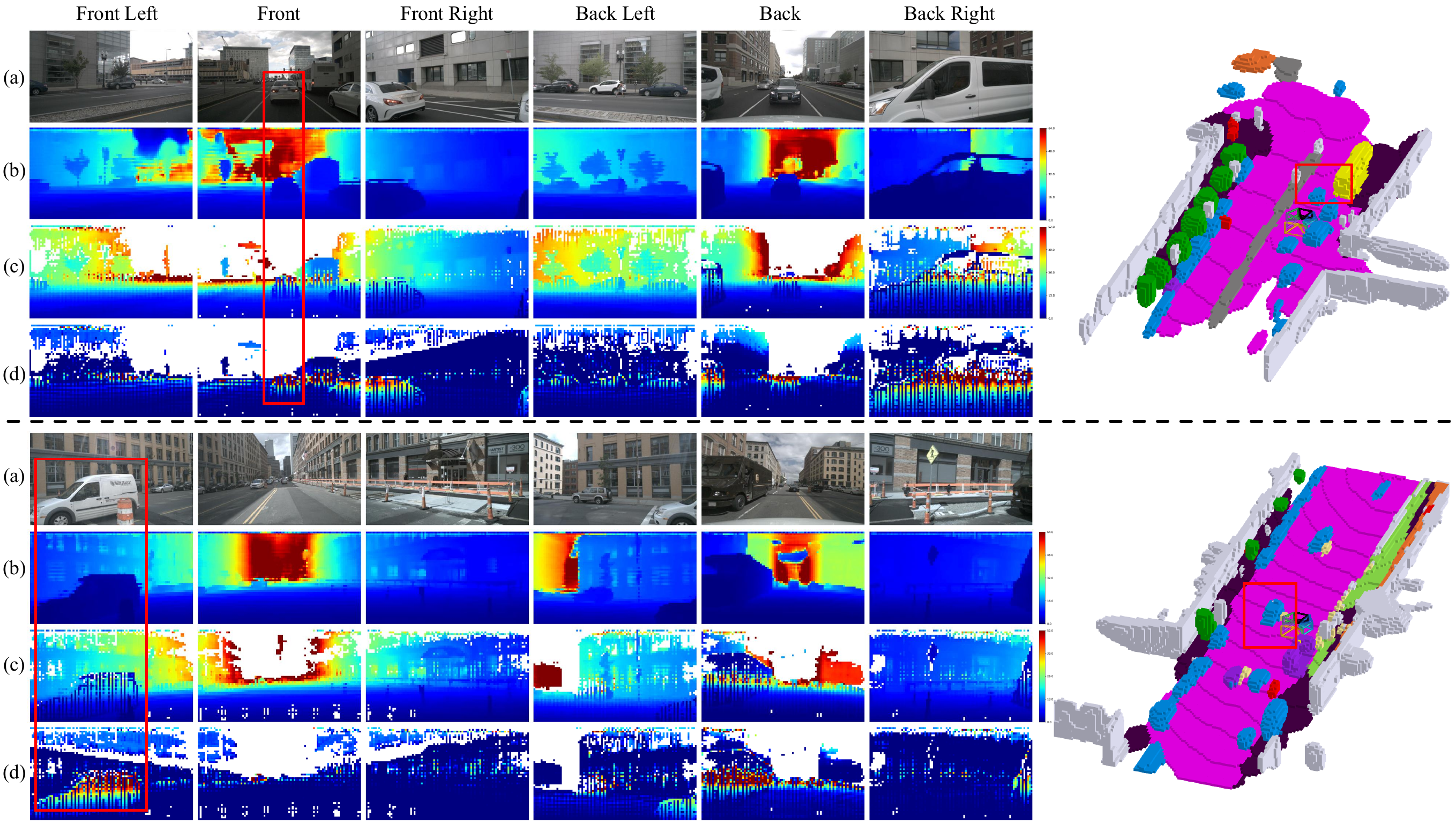}
\caption{\textbf{Occlusion-aware Gaussian positioning on SurroundOcc-nuScenes.} Results are produced by the jointly trained cross-scene OccAnyScene model. In the left panel, the six columns correspond to the six camera views, while the rows show (a) input images, (b) predicted surface depth, (c) the depth of the closest-to-camera Gaussian among the \(K=3\) Gaussians decoded from each pixel frustum, and (d) the corresponding surface-relative depth increment. Row (c) is obtained by adding (d) to (b). The right panel shows the final semantic occupancy prediction. Red boxes highlight representative foreground occlusions, where spatially alternating large increments displace Gaussians behind the visible surfaces.}
\label{fig:vis_nuscenes_depth}
\vspace{-2mm}
\end{figure*}

\begin{figure}[!t]
\centering
\vspace{-2mm}
\includegraphics[width=\linewidth]{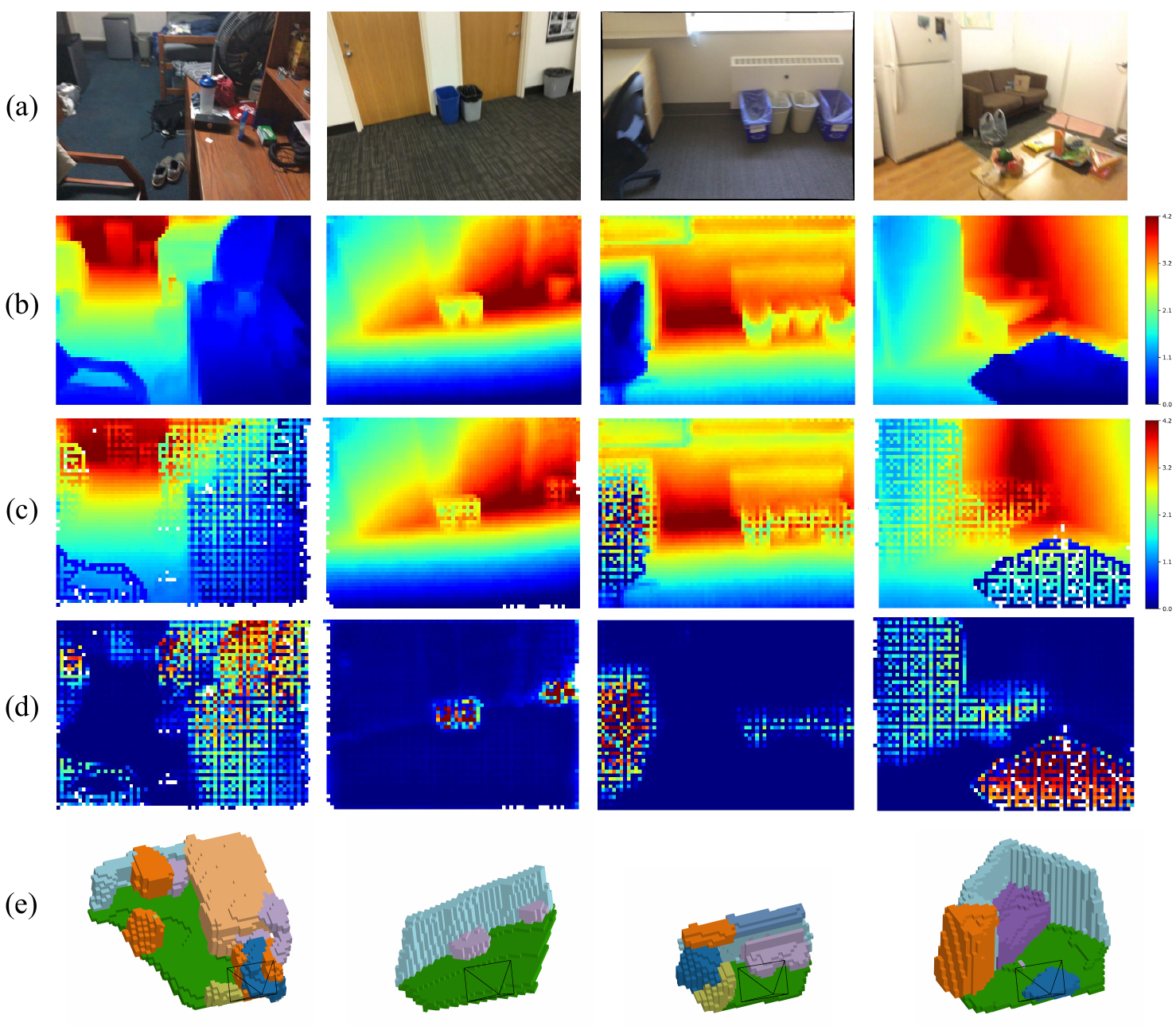}
\caption{\textbf{Occlusion-aware Gaussian positioning on Occ-ScanNet.} The rows show (a) the input image, (b) predicted surface depth, (c) the depth of the Gaussians decoded from each pixel frustum, (d) the surface-relative depth increment, and (e) the final semantic occupancy prediction. }
\label{fig:vis_scannet_depth}
\vspace{-8mm}
\end{figure}

Figures~\ref{fig:vis_nuscenes_depth} and~\ref{fig:vis_scannet_depth} visualize how the cross-scene OccAnyScene model positions Gaussians beyond visible surfaces. To obtain a single Gaussian-depth map for each view, we select the closest-to-camera Gaussian at each feature pixel, denoted by \(k^\star=\arg\min_k d_{p,k}\). Rows (b), (c), and (d) then visualize the predicted surface depth \(d_p^{\mathrm{surf}}\), the resulting Gaussian depth \(d_{p,k^\star}=d_p^{\mathrm{surf}}+\Delta d_{p,k^\star}\), and its surface-relative depth increment \(\Delta d_{p,k^\star}\), respectively.

In directly visible regions, the Gaussian depth remains close to the predicted surface depth, preserving the observed geometry. In foreground-occlusion regions highlighted by the red boxes, alternating large increments in (d) move some Gaussians behind the visible surface while neighboring Gaussians remain near it. The resulting Gaussian-depth maps in (c) therefore interleave foreground and inferred background depths, producing a transparency-like pattern. This behavior appears in both datasets and is particularly evident on Occ-ScanNet, where foreground-object regions in (d) exhibit mosaic-like high-increment patterns. These patterns suggest that the model uses contextual cues to identify likely occluders and collectively represent visible and occluded scene contents through neighboring pixels.

\section{Limitations}
Our current cross-scene evaluation is limited to Occ-ScanNet and SurroundOcc-nuScenes. Although these datasets cover substantially different indoor and outdoor settings, the experiments primarily demonstrate joint learning across two heterogeneous occupancy protocols rather than generalization to arbitrary unseen scenes. Extending OccAnyScene toward truly general-purpose occupancy prediction will require training on more diverse datasets spanning additional environments, sensing configurations, spatial scales, and semantic taxonomies.

The pixel-frustum-centered representation also has an inherent coverage limitation, as pixel frusta span only regions within the fields of view of the input cameras. Consequently, the pixel-frustum-derived Gaussians cannot directly represent the gaps between adjacent cameras in SurroundOcc-nuScenes. To cover these regions, we place a small set of spatial queries in the camera-unobserved areas and initialize their features as learnable embeddings. These queries interact through cross-attention with a pooled scene feature obtained by globally pooling the image tokens from all six views, and are decoded into supplementary Gaussian primitives. Therefore, the occupancy predictions within the inter-camera gaps shown in our nuScenes qualitative results are produced by these supplementary Gaussians rather than the pixel-frustum-derived Gaussians. Since these queries cover only a small fraction of the target space, they have little effect on the aggregate evaluation metrics. Extending occupancy prediction to substantially larger regions outside the camera fields of view would require more effective scene-level completion mechanisms.
\raggedbottom
\section{Conclusion}
We introduced Cross-Scene 3D Semantic Occupancy Prediction, where differences in camera geometry, metric range, voxel resolution, and semantic taxonomy make existing scene-specific representations difficult to optimize within one model. OccAnyScene addresses this problem by treating each pixel frustum as the basic unit for constructing a continuous Gaussian representation. It first aggregates geometric, camera, and contextual cues through Pixel-Aligned Frustum Feature Aggregation, and then adaptively parameterizes Gaussian positions and scales using Frustum-Parameterized Gaussian Construction. Experiments on Occ-ScanNet and SurroundOcc-nuScenes show that one cross-scene model achieves performance comparable to separately trained scene-specific models while maintaining high efficiency, demonstrating the effectiveness of OccAnyScene for unified indoor and outdoor occupancy prediction.

\clearpage
\flushbottom

{
    \small
    \bibliographystyle{ieeenat_fullname}
    \bibliography{reference}
}

\end{document}